\documentclass[letterpaper, 10 pt, conference]{ieeeconf}  

\usepackage{mathtools}
\usepackage{amsmath}
\usepackage{amssymb}
\usepackage{url}
\usepackage{graphicx}

\usepackage{color}
 
\usepackage{caption}
\usepackage{subcaption}
\usepackage{booktabs}
\usepackage{threeparttable}
\usepackage{multirow}

\usepackage[noadjust]{cite}

\usepackage{bbding}

\usepackage{bm} 

\let\labelindent\relax 
\usepackage[inline]{enumitem}

\IEEEoverridecommandlockouts                              

\title{\LARGE \bf
Traj-LO: In Defense of LiDAR-Only Odometry Using an Effective Continuous-Time Trajectory
}

\author{Xin~Zheng and
        Jianke~Zhu
\thanks{Xin~Zheng and Jianke~Zhu are with the College of Computer Science, Zhejiang University, Hangzhou, China, 310027. \protect\\ 
E-mail: \{xinzheng,jkzhu\}@zju.edu.cn.}
\thanks{Jianke Zhu is the Corresponding Author.}}

\begin{document}

\maketitle
\thispagestyle{empty}
\pagestyle{empty}

%
\begin{abstract}
LiDAR Odometry is an essential component in many robotic applications. Unlike the mainstreamed approaches that focus on improving the accuracy by the additional inertial sensors, this letter explores the capability of LiDAR-only odometry through a continuous-time perspective. Firstly, the measurements of LiDAR are regarded as streaming points continuously captured at high frequency. Secondly, the LiDAR movement is parameterized by a simple yet effective continuous-time trajectory. Therefore, our proposed Traj-LO approach tries to recover the spatial-temporal consistent movement of LiDAR by tightly coupling the geometric information from LiDAR points and kinematic constraints from trajectory smoothness. This framework is generalized for different kinds of LiDAR as well as multi-LiDAR systems. Extensive experiments on the public datasets demonstrate the robustness and effectiveness of our proposed LiDAR-only approach, even in scenarios where the kinematic state exceeds the IMU's measuring range. Our implementation is open-sourced on GitHub.

\end{abstract}

\section{INTRODUCTION}

LiDAR Odometry (LO) plays a crucial role in robot navigation and path planning within GPS-denied environments~\cite{cadena2016past} due to its exceptional range sensing ability. Over the past decade, increasing research efforts have been devoted to the aspects like feature selection~\cite{deschaud2018imls,pan2021mulls}, map representation~\cite{behley2018efficient,zheng2021efficient}, and sensor fusion~\cite{shan2020lio,xu2022fast} to enhance the precision, efficiency, and robustness of LO systems. 

\begin{figure}[t]
	\centering
	\includegraphics[width=0.5\textwidth]{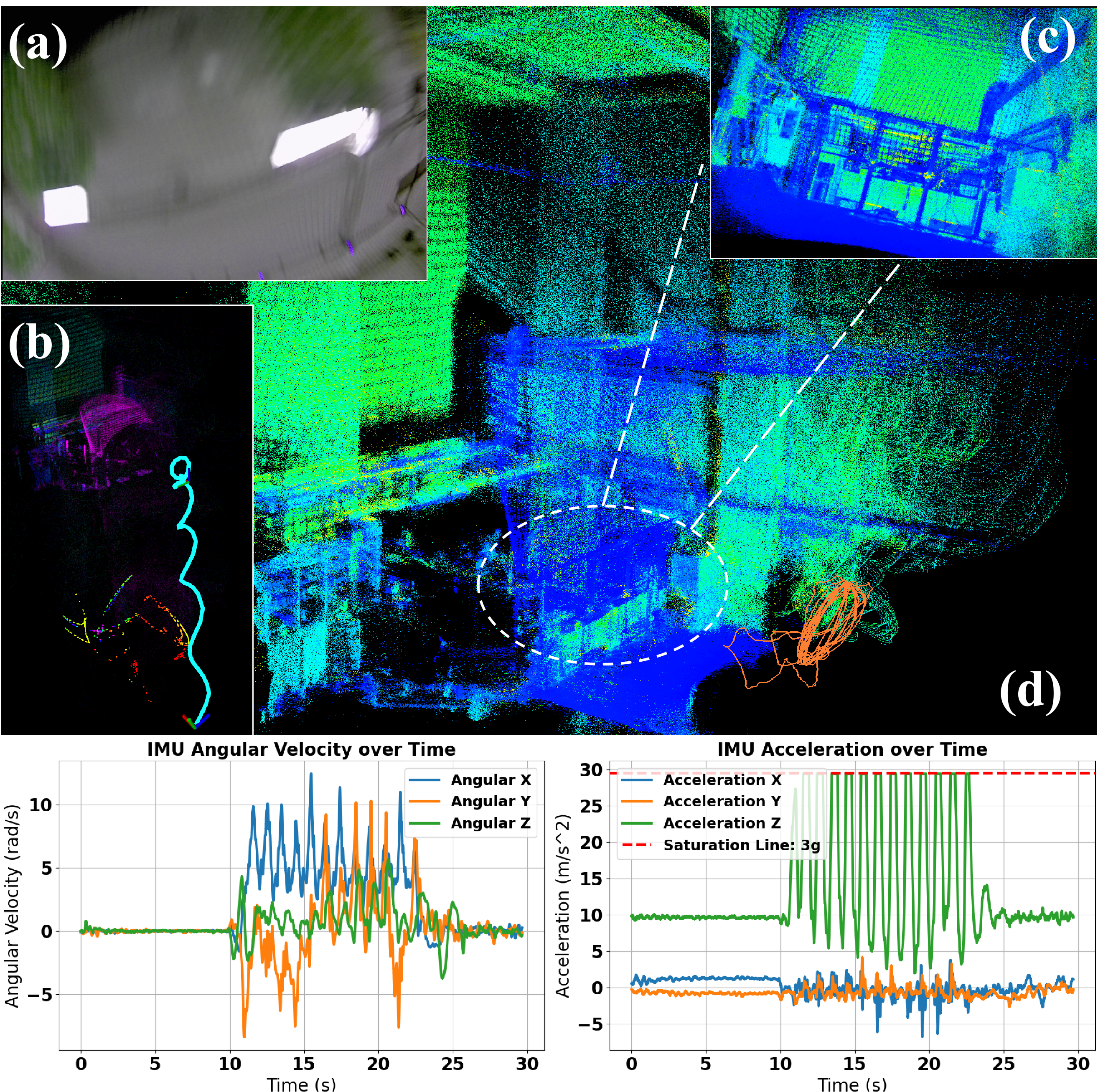}
	\caption{\small \textit{The upper figure shows the mapping result in case that acceleration exceeds IMU measuring range 3G and (a) the camera sensor is totally blurred in fast motion. (b)FAST-LIO~\cite{xu2022fast} drift significantly (c) while our method is still valid only using LiDAR.}}
	\label{fig:extreme}
	\vspace{-0.3in}
\end{figure}

Currently, tightly-coupled LiDAR-inertial Odometry~\cite{shan2020lio, xu2022fast} (LIO) leverages information from Inertial Measurement Units (IMUs), which is widely regarded as the optimal solution in practical applications~\cite{helmberger2022hilti, nguyen2022ntu}. Comparing to the existing dominant LO methods~\cite{zhang2014loam, zheng2021efficient, wang2021f, dellenbach2022ct, vizzo2023ral}, LIO approach offers advantages on two folds. One is that the high-frequency propagated IMU states contribute to more accurate motion compensation. Another advantage is  that IMUs enhance system robustness by directly enforcing kinematic constraints, either through pre-integration~\cite{forster2015manifold} or Kalman Filters~\cite{xu2022fast}. Unfortunately, a stable LIO heavily relies on the accurate calibration of IMU parameters, and IMU data are sensitive to temperature variations and mechanical shocks. Furthermore, recent research~\cite{he2023point} finds that even the state-of-the-art LIO systems~\cite{shan2020lio,xu2022fast} struggle to handle scenarios depicted in Fig.~\ref{fig:extreme}, where the kinematic state surpasses the IMU measurement range. These challenges motivate us to investigate the potential of LiDAR-only approach in the extreme situations that are usually considered insurmountable.

Indeed, LiDAR is a streaming sensor with continuous movements in order to capture scene structures. The standard LO paradigm~\cite{zhang2014loam} typically represents the trajectory as a sequence of discrete-time poses, which severs the inherent connection between geometry and kinematics. Ideally, the geometry of continuous scene structures should reflect the spatial-temporal movement of LiDAR, while kinematics has motion-coherent constraints~\cite{myronenko2010point} for consecutive observed points. Hence, our proposed odometry approach adopts a continuous-time trajectory to tightly couple geometric information and kinematic prior hiding behind streaming points.

The key of continuous-time trajectory is the ability to query the reference LiDAR pose of each measured point by its corresponding timestamp. Therefore, the previously neglected temporal information can be employed to iteratively adjust the point position without the need of motion compensation. This continuous-time registration technique~\cite{dellenbach2022ct, zheng2023ectlo} ensures not only spatial consistency but also temporal coherence. Moreover, the continuous-time trajectory introduces an indirect kinematic constraint through its smoothness, which is vital for preventing divergence during registration. To overcome the inefficiency of current methods~\cite{wu2022picking,quenzel2021real,lv2021clins}, our trajectory is composed of multiple linear segments. This enables us to approximate continuous movement via the simplest linear interpolation within each segment. Furthermore, the trajectory is represented in the compact SE(3) parameterization, which greatly facilitates the derivation of analytic Jacobians. Additionally, the continuous-time trajectory inherently accommodates the fusion of multiple asynchronous LiDARs into a unified optimization framework, which presents a natural advantage in scenarios with multiple LiDAR sensors.

In summary, the main contributions of this paper are: 1) the LiDAR-only odometry approach with a simple yet effective continuous-time trajectory, adaptable to various LiDAR types and multi-LiDAR systems. The implementation of the framework will be open-source to benefit the community; 2) spatiotemporal consistent registration by leveraging previously overlooked temporal information from millions of points and the inherent smoothness of the continuous trajectory; 3) extensive experiments on diverse public datasets to demonstrate that our LiDAR-only method matches the performance of state-of-the-art LIO approaches~\cite{shan2020lio,xu2022fast} and even surpasses them in extreme scenarios.

\section{RELATED WORKS}\label{sec:rel}
\subsection{LO and LIO Using Discrete-time Trajectory} 
LOAM~\cite{zhang2014loam} is the seminal work on LiDAR odometry, which divided the points into line and planar features based on the roughness of each point on its respective ring. Then, a modified Iterative Closest Point (ICP) algorithm~\cite{besl1992method} incorporating both point-to-line and point-to-plane metrics was utilized to estimate poses at a scan rate of 10~Hz. Many subsequent works~\cite{deschaud2018imls, pan2021mulls, wang2021f, zheng2021efficient, vizzo2023ral} aim to enhance the performance on the KITTI Odometry benchmark~\cite{geiger2012we}. Notably, F-LOAM~\cite{wang2021f} introduced a lightweight scheme with efficient optimization. Recently, KISS-ICP~\cite{vizzo2023ral} achieved the remarkable performance by adopting simple point-to-point ICP, complemented by a constant velocity model to mitigate motion distortion~\cite{helmberger2022hilti, nguyen2022ntu}. However, existing LiDAR-only approaches struggle with the aggressive motions.

Practically, LIO aided by additional IMU measurements is recognized as a more robust solution. Shan et al.~\cite{shan2020lio} formulated LIO by a factor graph framework to integrate various constraints, including loop closure, wheel odometry, and GPS, into the LIO system. Xu et al.~\cite{xu2022fast} fused LiDAR and IMU measurements within an iterative error-state Kalman Filter framework, which corrected distorted points through back-propagation and directly registered raw points onto an incremental kd-tree map for fast neighborhood searching.

The above methods represent LiDAR movement by a series of discrete-time poses. The points collected within the period of consecutive poses have to be transformed into the position at a specific timestamp. This is called motion compensation, which is usually done before registration.

\subsection{LO and LIO Using Continuous-time Trajectory}
Indeed, the LiDAR points are streaming data captured continuously. Previous methods of computing rigid transformation between two discrete-time poses suffer from motion distortions, where the reference LiDAR center of each measured point is not exactly the same as body movement. In general, the continuous-time trajectory~\cite{furgale2012continuous} has the ability to query any state of sensors' spatial-temporal movement at any given timestamp. The challenge is how to approximate the entire trajectory with finite states while preserving accuracy and ensuring efficiency.

One major class is based on Gaussian Processes (GP)~\cite{tong2013gaussian, barfoot2014batch, dong2018sparse, wu2022picking}. Tong et al.~\cite{tong2013gaussian} employed GP to model the continuous movement in SLAM, where Gauss-Netween method is used for state updating. Barfoot et al.~\cite{barfoot2014batch} explored the sparsity by incorporating time-varying prior, which makes GP regression more efficient.
Anderson~\cite{anderson2015full} conducted batch continuous-time trajectory estimation in SE(3) space, while Dong~\cite{dong2018sparse} extended sparse GP regression to the general Lie Group. More recently, Wu et al.~\cite{wu2022picking} incorporated Doppler velocity measurements from FMCW LiDAR into GP-based trajectory estimation. However, it only achieves 2~Hz, which is the common limitation of GP methods.

Another solution is the parametric trajectory, especially B-spline~\cite{sommer2020efficient}. Quenzel and Behnke introduced a surfel-based B-spline approach, also known as MARS~\cite{quenzel2021real}, which validated across both regular driving and aerial scenarios. CLINS~\cite{lv2021clins} integrated IMU data into trajectory estimation, although it struggled to meet real-time requirements. Its successor, CLIC~\cite{lv2023continuous}, extended the B-spline representation to the LiDAR-inertial-camera system.

In contrast to B-splines which are weighted sums of basis functions, linear interpolation offers a more straightforward parametric approach to describing trajectories. However, linear interpolation at a low-frequency scan rate becomes inaccurate when facing aggressive motion. CT-ICP~\cite{dellenbach2022ct} simply addressed this issue by attributing the errors to scan discontinuities and mitigating them through the additional motion constraints. It has proven effective in ordinary driving scenarios but is unreliable for handheld devices and aerial vehicles. Therefore, other approaches such as Zebedee~\cite{bosse2012zebedee}, ElasticSLAM~\cite{park2021elasticity}, and Wildcat~\cite{ramezani2022wildcat} adopted a series of high-frequency control poses with smaller temporal intervals to reduce linear errors caused by irregular motions. Nonetheless, these methods heavily rely on IMU measurements for robust estimation and also require a tedious cubic B-spline fitting procedure to refine the trajectory.

\section{METHODOLOGY}
\subsection{Preliminary}

To facilitate subsequent derivations, we firstly clarify the notations in the following. We define the continuous-time trajectory as $\mathbf{T}(t)$ over the interval $[t_{0},t_{K})$. For a specific pose at timestamp $t$, we denote it as $\mathbf{T}_{t}$. All poses are represented in world coordinates, where the world frame is the starting location of LiDAR. The concrete LiDAR pose $ \mathbf{T}= \begin{bmatrix}
  \mathbf{R}&\mathbf{t}  \\
  0& 1
\end{bmatrix}$ lies on $\mathrm{SE(3)}$ manifold~\cite{sola2018micro}, where $\mathbf{t} \in \mathbb{R}^{3}$ represents the translation and $\mathbf{R} \in \mathrm{SO(3)}$ denotes the rotation. The introduction of $\mathrm{SE(3)}$ enables to formulate the 3D rigid transformation into matrix multiplication. 

As in~\cite{sola2018micro}, we employ the right plus $\oplus$ and minus $\ominus $ to establish a connection between the vector $\boldsymbol{\tau} \in \mathbb{R}^{6}$ in tangent space $\mathfrak{se}(3) $ and transformation matrix $\mathbf{T}\in\mathrm {SE}(3) $. Incrementing a transformation matrix by a vector is denoted as$\mathbf{T}\oplus \boldsymbol{\tau}=\mathbf{T}\mathrm{Exp}(\boldsymbol{\tau})$, while difference between two transformation matrix is given by $\mathbf{T}_{1}\ominus \mathbf{T}_{2}=\mathrm{Log}(\mathbf{T}_{2}^{-1}\mathbf{T}_{1})$. $\mathrm{Exp}:\mathbb{R}^{6}\to \mathrm{SE}(3) $ and $\mathrm{Log}: \mathrm{SE}(3) \to\mathbb{R}^{6}$ are the mapping functions.

\subsection{Continuous-Time Trajectory Parameterization}\label{sec:traj}
\begin{figure}[t]
	\centering
\includegraphics[width=0.48\textwidth]{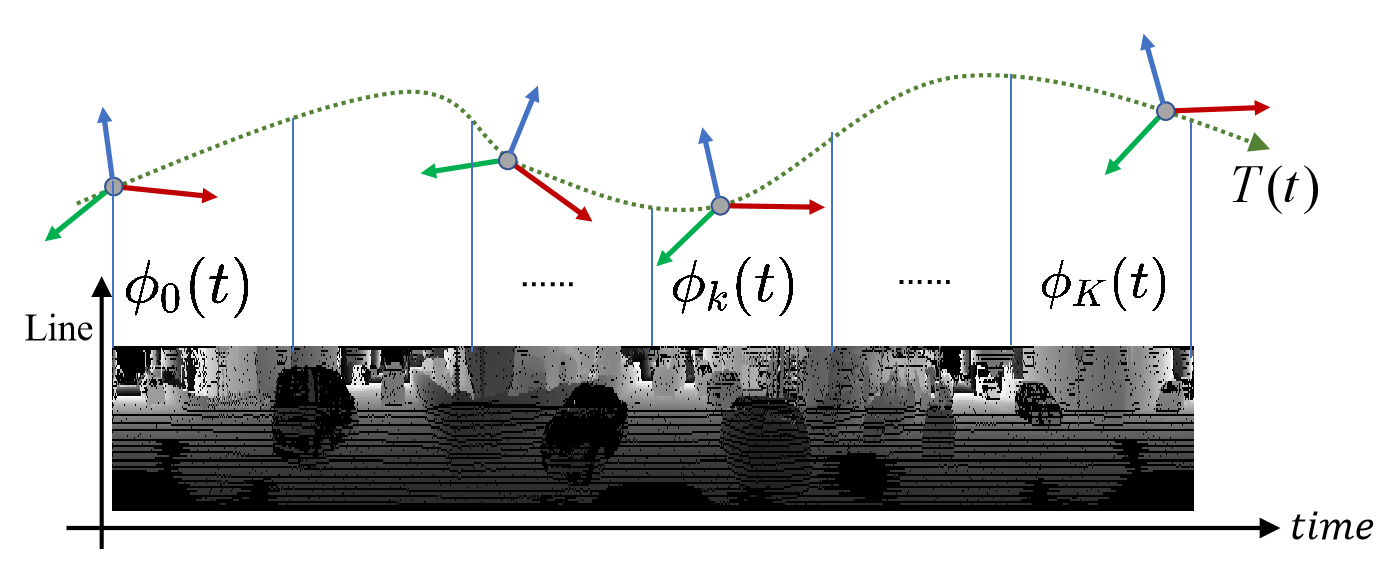}
	\caption{\small \textit{Trajectory $\mathbf{T}(t)$ shows the continuous movement of LiDAR over interval $[t_{0},t_{K})$. The bottom is a range image~\cite{zheng2021efficient} of point cloud collected by a 64-line LiDAR in a $360^{\circ}$ circular motion. Only points in the same column are measured simultaneously}}
	\label{fig:traj}
	\vspace{-0.2in}
\end{figure}

In this letter, we introduce the continuous-time trajectory parameterization which serves as the foundation. As depicted in Fig.~\ref{fig:traj}, LiDAR is a streaming sensor typically with continuous movement. Points in different columns of the range image are indeed collected at distinct LiDAR poses. By employing a continuous-time trajectory, we can perform registration using the precise LiDAR pose without requiring prior motion compensation. The challenge is how to represent continuous-time trajectories accurately and efficiently.

To tackle this issue, we propose a straightforward yet effective scheme. Specifically, the simplest linear interpolation~\cite{dellenbach2022ct} of the trajectory over the temporal window $[t_{0},t_{K}) $ can be modeled by the beginning pose $\mathbf{T}_{0}$ and end pose $\mathbf{T}_{K}$. Unfortunately, linear approximation fails to obtain satisfactory results in case of aggressive motions such as handheld devices or aerial vehicles. Indeed, the body's movement becomes more linear as $t_{K}-t_{0} \to 0$. However, a short interval implies fewer accumulated points, which may potentially lead to under-constrained situations during the registration process.

To account for rapid motion, we divide the temporal window into $K$ equidistant small-resolution segments $  \left \{ [t_{k-1},t_{k}) \right \}^{K}_{k=1}$. In each segment $[t_{k-1},t_{k})$, the movement is still parameterized with two control poses, the initial $\mathbf{T}_{k-1}$ and final $\mathbf{T}_{k}$. We represent the continuous-time trajectory over this segment as function $\phi_{k}(t)$. For a measured point with timestamp $t_{i}$ in this segment, we retrieve its associated reference LiDAR pose as $\mathbf{T}_{t_{i}}=\phi_{k}(t_{i})$
\begin{equation}
\label{equ:interplotation}
\phi_{k}(t_{i}) =\mathbf{T}_{k-1}\oplus(\alpha_{i} \boldsymbol{\tau}_{k} ),
\end{equation}
where $\alpha_{i}=(t_{i}-t_{k-1})/\Delta t_{k}$, and $\boldsymbol{\tau}_{k} =
    \mathbf{T}_{k}\ominus\mathbf{T}_{k-1}$. The interval length of each segment $\Delta t_{k}=t_{k}-t_{k-1} $ is a hyperparameter that depends on the motion profile. Then, the entire trajectory $ \mathbf{T}(t)$ over temporal window $[t_{0},t_{K})$ consist of $K$ piecewise functions
$\left \{  \phi_{1}(t),   \dots,\phi_{k}(t),\dots,   \phi_{K}(t)\right \}$. 
The trajectory $\mathbf{T}(t)$ during the $k$-th temporal interval $[t_{k-1},t_{k})$ corresponds to the $k$-th segment function, where $\mathbf{T}(t)=\phi_{k}(t)$. Thus, we can narrow the interval to hold an accurate linear approximation in each segment in case of aggressive motion. Meanwhile, the abundant measurements collected over the entire window provide the necessary geometric constraints for robust trajectory estimation.

Finally, the continuous-time trajectory $\mathbf{T}(t)$ within the temporal window $[t_{0},t_{K})$ actually depends on $K+1$ control poses
$\left \{  \mathbf{T}_{0},   \dots,\mathbf{T}_{k},\dots,   \mathbf{T}_{K}\right \}$. Unlike the discrete-time trajectory that only indicates the LiDAR position at $K+1$ moments, our proposed approach employs these $K+1$ control poses to model the entire continuous-time trajectory. At the first glance, our piecewise continuous-time trajectory is quite simple to implement. By properly handling three parts in Sec.~\ref{sec:LO}, LiDAR-only odometry using such parameterization performs on par with the state-of-the-art LIO methods~\cite{xu2022fast,shan2020lio}.

\subsection{LiDAR Odometry Using Continuous-Time Trajectory}\label{sec:LO}
\begin{figure}[t]
	\centering
	\includegraphics[width=0.5\textwidth]{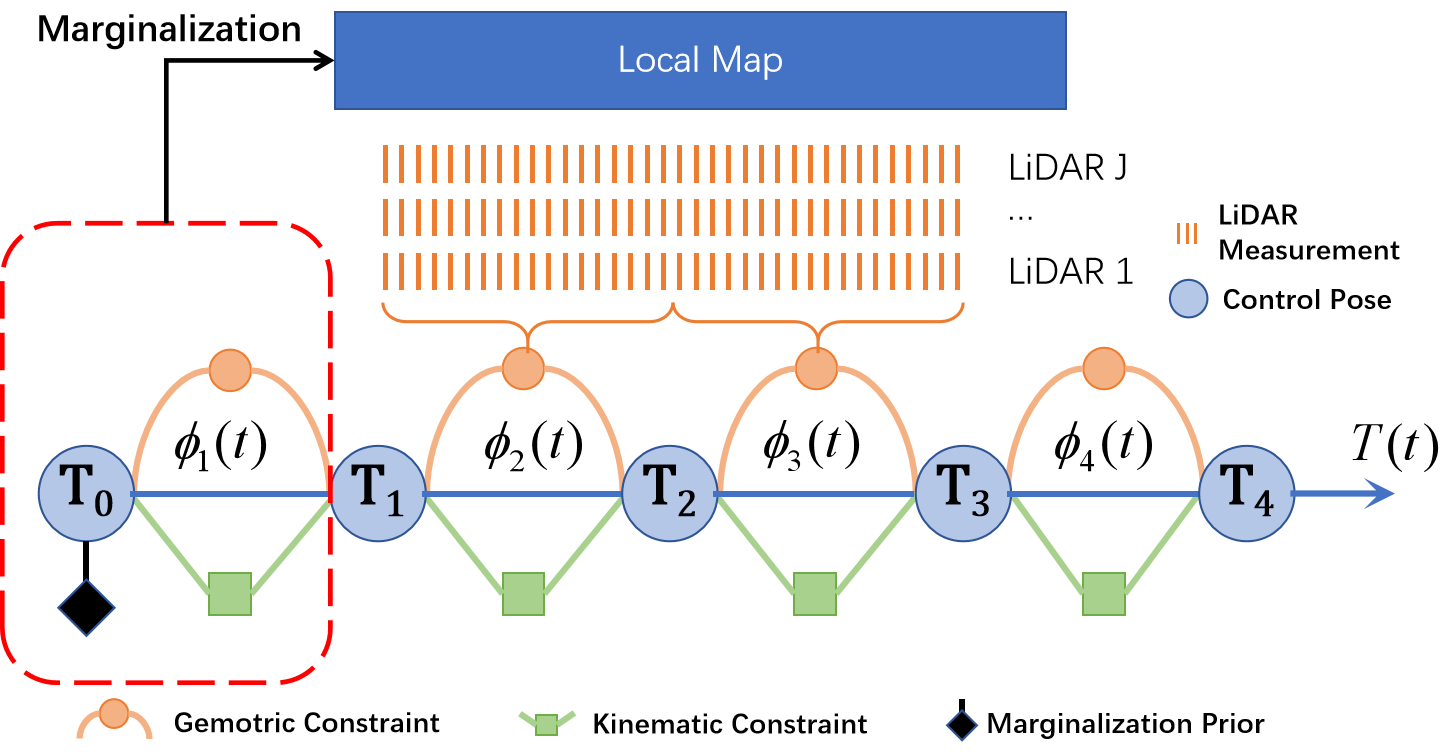}
	\caption{\small \textit{A trajectory $\mathbf{T}(t)$ consisted with 4 segments, where each segment $\phi_{k}(t)$ is modeled by linear interpolation in Sec.~\ref{sec:traj}. There are three kinds of constraints, including geometry, kinematics, and marginalization.}}
	\label{fig:pipeline}
	\vspace{-0.2in}
\end{figure}

We present the LiDAR odometry approach using the presented trajectory parameterization. Fig.~\ref{fig:pipeline} illustrates the whole pipeline. To facilitate the real-time requirement, our method works in a sliding window fashion, which consists of three main components, including continuous-time registration, kinematic constraint, and marginalization.

To derive the probabilistic formulation of LiDAR odometry, we introduce the control input $\mathbf{u}(t)$ and map $\mathcal{M}\doteq  \left \{ \mathbf{q}_{m}  \right \}^{M}_{m=1} $. The continuous-time variable $\mathbf{u}(t)$ affects the system's dynamics, while $\mathcal{M}$ is constructed by LiDAR points before the current temporal window. Each map point $\mathbf{q}_{m}$ is defined in the world coordinates, which is discussed in Sec.~\ref{sec:map}. The estimated trajectory $\mathbf{T}(t)$ over interval $[t_{0},t_{K})$ depends on a series of LiDAR measurements $\mathcal{Z}$. Our target is to seek the maximum joint posterior distribution $p(\mathbf{T}(t) \mid \mathbf{u}(t), \mathcal{M},\mathcal{Z} )$ over this interval.

Since the control input $\mathbf{u}(t)$ does not influence the LiDAR measurements for the given $\mathbf{T}(t)$, we rewrite the posterior using Bayes' rule as follows:
$$
\label{equ:bays}
p(\mathbf{T}(t) \mid \mathbf{u}(t), \mathcal{M},\mathcal{Z} )\propto p(\mathbf{ \mathcal{Z} }\mid \mathbf{T}(t),\mathbf{ \mathcal{M} })p(\mathbf{T}(t)\mid \mathbf{u}(t)).
$$

\subsubsection{Continuous-Time Geometric Constraint}\label{sec:geometry}
The first part $p(\mathbf{ \mathcal{Z} }\mid \mathbf{T}(t),\mathbf{ \mathcal{M} })$ encodes the geometric relationship. Instead of just focusing on a specific LiDAR, our method is designed to accommodate various types of LiDARs, such as multi-line spinning and non-repetitive LiDARs, as well as different configurations of LiDAR systems, whether single or multiple. Given the potential involvement of multiple LiDARs in our system, we assume the continuous trajectory $\mathbf{T}(t)$ rigidly attached to a reference coordinate system that typically corresponds to the first LiDAR.

According to our trajectory parameterization, we organize all measurements $ \mathcal{Z}$ into $K$ sets, denoted as $\mathcal{Z}\doteq \left \{ \mathcal{Z}_{k}  \right \}_{k=1}^{K} $. Each set $\mathcal{Z}_{k}$ includes measurements collected within the interval $[t_{k-1},t_{k})$ from $J$ LiDARs. The specific form is
\begin{equation}
  \mathcal{Z}_{k}\doteq  \left \{ \left \{ {_{k}\mathbf{z}_{i}^{j}}  \right \}_{i=1}^{N_{k}^{j}}  \right \}_{j=1}^{J}, \quad 
  _{k}\mathbf{z}_{i}^{j}=\left \{  {_{k}\mathbf{p}_{i}^{j}},{_{k}t_{i}^{j}} \right \}    .
\end{equation}
where $N_{k}^{j}$ is the point number measured by $j$-th LiDAR in $k$-th segment. Note that each measurement $_{k}\mathbf{z}_{i}^{j}$ consists of a pair, i.e., the point's 3D position $_{k}\mathbf{p}_{i}^{j}=({_{k}x_{i}^{j}} ,{_{k}y_{i}^{j}} ,{_{k}z_{i}^{j} })$ and its corresponding timestamp ${_{k}t_{i}^{j}} $. Additionally, the extrinsic calibration parameter for each LiDAR relative to the reference coordinate system is represented by $\mathbf{T}^{B}_{L_{j}} $. These parameters are pre-calibrated and remained to be fixed throughout the optimization process. 

Assuming that all measurements are independent, $p(\mathbf{ \mathcal{Z} }\mid \mathbf{T}(t),\mathbf{ \mathcal{M} })$ can be factorized as follows
\begin{equation}
    {\textstyle \prod_{k=1}^{K}} {\textstyle \prod_{j=1}^{J}} {\textstyle \prod_{i=1}^{N_{k}^{j}}} p(\left \{ _{k}\mathbf{p}_{i}^{j},{_{k}t_{i}^{j}}   \right \} \mid \mathbf{T}({_{k}t_{i}^{j}} ),\mathbf{ \mathcal{M} }).
\end{equation}
Existing LOs~\cite{wang2021f,vizzo2023ral} that utilize the discrete-time trajectories often assume that all points are measured simultaneously at a specific timestamp. Thus, the accuracy of these systems is highly dependent on the quality of motion compensation.

In contrast, our approach implements a spatial-temporal consistent registration. We not only consider the 3D point positions but also pick up the neglected temporal information, which allows for iterative updates of the point positions. Since we can query the reference LiDAR pose for each point using the continuous-time trajectory, it is unnecessary to take consideration of motion distortion. Moreover, we use the point-to-plane criteria for registration without selecting feature. This makes our approach to be effective for different types of LiDAR. The geometric error of ${_{k}\mathbf{z}_{i}^{j}}$ is 
\begin{equation}
    \mathbf{e}_{_{k}\mathbf{z}_{i}^{j}}={_{k}\mathbf{n}_{i}^{j}}^{\top } (\phi_{k}({_{k}t_{i}^{j}})\cdot \mathbf{T}^{B}_{L_{j}} \cdot {_{k}\mathbf{p}_{i}^{j}}-{_{k}\mathbf{q}_{i}^{j}}  ).
\end{equation}
where $\phi_{k}({_{k}t_{i}^{j}})$ is the pose of reference coordinate system computed by Equ.~\ref{equ:interplotation}. ${_{k}\mathbf{q}_{i}^{j}} $ is the nearest neighbor of ${_{k}\mathbf{p}_{i}^{j}}$ in the map $\mathcal{M}$. The normal vector ${_{k}\mathbf{n}_{i}^{j}}$ is computed by Principal Component Analysis choosing five closet points in $\mathcal{M}$ like~\cite{xu2022fast,dellenbach2022ct}. 

Ideally, the point-to-plane observation should be close to zero. We assume that $\mathbf{e}_{_{k}\mathbf{z}_{i}^{j}} \sim \mathcal{N}(\mathbf{0} ,\mathbf{Q}_{r} ) $ accounts for the LiDAR measurement error. Therefore, we have:
\begin{equation}
    p(\left \{ _{k}\mathbf{p}_{i}^{j},{_{k}t_{i}^{j}}   \right \} \mid \mathbf{T}({_{k}t_{i}^{j}}),\mathbf{ \mathcal{M} })
=\mathcal{N}(\mathbf{e}_{_{k}\mathbf{z}_{i}^{j}},\mathbf{Q}_{r}) 
\end{equation}

\subsubsection{Trajectory Smoothness Constraint}~\label{sec:kinematic}
The second part $p(\mathbf{T}(t)\mid \mathbf{u}(t))$ encodes the kinematic model. It is typically utilized in LIOs~\cite{shan2020lio,xu2022fast} but is often neglected in LO. This is because the control input $\mathbf{u}(t)$ can directly connect to the IMU data. Indeed, kinematics is an inherent property of the system, even without the assistance of IMU. Here we propose an indirect motion constraint from trajectory smoothness, which is vital for the convergence of continuous-time registration only depending on LiDAR input.

Given that our trajectory parameterization has divided the temporal window into several small intervals, the velocity within each segment is approximately constant. Then, we define the generalized velocity $\varpi_{k} (t)$ in $\mathrm{SE}(3)$ over the interval $[t_{k-1},t_{k})$ as $  \varpi_{k} (t)=( \mathbf{T}_{k}\ominus\mathbf{T}_{k-1})/ \Delta_{k}$. In the physical world, a continuous trajectory should exhibit smoothness, where the velocity between consecutive segments tends to be coherent. Thus, we introduce a pseudo-velocity measurement $\check{\varpi_{k}} = ( \check{\mathbf{T}} _{k-1} \ominus\check{\mathbf{T}} _{k-2})/ \Delta_{k-1}$ from previous segment and  ensure $\varpi_{k} (t)-\check{\varpi_{k}}$ to be close to zero. $\check{\mathbf{T}} _{k-1}$ and $\check{\mathbf{T}} _{k-2}$ are fixed values obtained after the convergence of the previous sliding window optimization, rather than variables in the current optimization window. Since the intervals of each segment are equidistant, we directly define the smoothness error through control poses as:
\begin{equation}
    \mathbf{e}_{v_{k}}= \mathbf{T}_{k}\ominus\mathbf{T}_{k-1}- \check{\mathbf{T}} _{k-1}\ominus\check{\mathbf{T}} _{k-2}.
\end{equation}
Let $\mathbf{e}_{v_{k}}\sim \mathcal{N}(0,\mathbf{Q}_{v} ) $ follows a zero-mean Gaussian distribution, $p(\mathbf{T}(t)\mid \mathbf{u}(t))$ over the entire trajectory is split into 

$$
    p(\mathbf{T}(t)\mid \mathbf{u}(t))= {\textstyle \prod_{k=1}^{K}}p(\phi_{k}(t),\mathbf{u}_{k}(t) )
$$
where $p(\phi_{k}(t),\mathbf{u}_{k}(t) )=\mathcal{N}(\mathbf{e}_{v_{k}} ,\mathbf{Q}_{v} )$.

\subsubsection{Joint Optimization and Marginalization}~\label{sec:marg}
Our target is to find the optimal continuous-time trajectory $\mathbf{T}(t)$ depending on $K+1$ control poses $\left \{ {\mathbf{T}_{k}} \right \}^{K}_{k=0}$ over the temporal window $[t_{0},t_{K})$. Thus, we minimize the negative logarithm of the posterior likelihood as below
\begin{equation}
    \left \{ {\mathbf{T}_{k}}^{*}  \right \}^{K}_{k=0}= \mathop{\mathrm{argmin}}_{\mathbf{T}(t) }(-\log (p(\mathbf{T}(t) \mid \mathbf{u}(t), \mathcal{M},\mathcal{Z} )) .
\end{equation}
Since all distributions are Gaussian, it can be transferred into a non-linear least squares problem, which is equal to minimize the following energy function
\begin{equation}
    \left \{ {\mathbf{T}_{k}}^{*}  \right \}^{K}_{k=0}=\mathop{\mathrm{argmin}}_{\mathbf{T}(t) }E_{reg}+E_{kine}+E_{marg} 
    \label{equ:traj},
\end{equation}
$$
    E_{reg}={\sum_{k=1}^{K}} \sum_{j=1}^{J}\sum_{i=1}^{N_{k}^{j}}\mathbf{e}_{_{k}\mathbf{z}_{i}^{j}}^{\top } \mathbf{Q}_{r}^{-1}\mathbf{e}_{_{k}\mathbf{z}_{i}^{j}},\: 
    E_{kine} ={\sum_{k=1}^{K}}\mathbf{e}_{v_{k}}^{\top}\mathbf{Q}_{v}^{-1}\mathbf{e}_{v_{k}}.
$$

We have provided the formulation for error terms $\mathbf{e}_{_{k}\mathbf{z}_{i}^{j}}$ and $\mathbf{e}_{v_{k}}$, while $E_{marg}$ represents the marginalization energy resulting from the sliding window optimization. To ensure real-time performance, we marginalize the oldest segment $\phi_{1}(t)$, once the window exceeds the predefined threshold. Instead of simply discarding measurements collected over $[t_{0},t_{1})$, we employ the marginalization priors to retain information that falls outside the next temporal window.

As described in~\cite{demmel2021square}, the computation of the marginalization priors $\mathbf{H}_{m}$ and $\mathbf{b}_{m}$ occurs after the optimization in the old window has converged. We derive $\mathbf{H}_{m}$ and $\mathbf{b}_{m}$ using Schur complement, considering only the energy terms that depend on the marginalized variables. For more detailed procedures, please refer to~\cite{demmel2021square}. The control poses in the new optimization window can be split into two parts $\mathbf{s}=\left \{ \mathbf{s}_{m}, \mathbf{s}_{n}\right \} $, where $\mathbf{s}_{m}$ is related to marginalized poses, and $\mathbf{s}_{n}$ is irrelevant. The marginalization energy is
$$
    E_{m}=\frac{1}{2}(\mathbf{s}_{m} \ominus \bar{\mathbf{s}}_{m})^{\top}\mathbf{H}_{m}(\mathbf{s}_{m} \ominus \bar{\mathbf{s}}_{m})+\mathbf{b}_{m}^{\top}(\mathbf{s}_{m} \ominus \bar{\mathbf{s}}_{m}),
$$
and only influences $\mathbf{s}_{m}$. $\bar{\mathbf{s}}_{m} $ is the fixed linearization point when applying Schur complement. In our presented trajectory representation, segment $\phi_{1}(t)$ depends on two control poses, $\mathbf{T}_{0}$ and $\mathbf{T}_{1}$. Since $\mathbf{T}_{1}$ is relevant to the next segment $\phi_{2}(t)$, only $\mathbf{T}_{0}$ will be removed during the marginalization. Besides, the energy term depends on $\mathbf{T}_{0}$, which is related to $\mathbf{T}_{1}$. Thus, $\mathbf{s}_{m}$ in new window includes the poses $\mathbf{T}_{1}$ from the old window. In addition, we apply the first-estimate Jacobians~\cite{huang2009first} to keep the consistency of the reduced system. For the poses connected to marginalization prior, we derived Jacobin at their fixed linearization point $\bar{\mathbf{s}}_{m}$, while the residuals are calculated at their current state $\mathbf{s}_{m}$.

\subsection{Analytic Jacobian Derivation}

We make use of Gauss-Netwon optimization to solve the nonlinear least square minimization in Equ.~\ref{equ:traj}. By the first-order Taylor expansion, the general error term around the linearized point $\mathbf{s}$ is approximated as $    \mathbf{e}(\mathbf{s}\oplus \boldsymbol{\xi})\simeq \mathbf{e}(\mathbf{s})+\frac{\partial \mathbf{e}}{\partial \mathbf{s}} \boldsymbol{\xi}$, where $\mathbf{s}$ is $K+1$ poses with its stacked increment vector $\boldsymbol{\xi} \in \mathbb{R}^{6(K+1)} $. $\partial \mathbf{e}/ \partial \mathbf{s}$ is the Jacobian. The optimal increment is found by solving the normal equation $\mathbf{H}\boldsymbol{\xi}=-\mathbf{b}$, where the Hessian matrix $\mathbf{H}=\sum \left ( \frac{\partial \mathbf{e}}{\partial \mathbf{s}} \right )^{\top }\mathbf{W}^{-1}\frac{\partial \mathbf{e}}{\partial \mathbf{s}}$ and $\mathbf{b}=\sum \left ( \frac{\partial \mathbf{e}}{\partial \mathbf{s}} \right )^{\top }\mathbf{W}^{-1}\mathbf{e}$. $\mathbf{W}$ is the related covariance. Each control pose of our continuous trajectory iteratively updates  as $\mathbf{T}_{k}\leftarrow  \mathbf{T}_{k}\oplus \boldsymbol{\xi}_{k}$, where $\mathbf{s}=\begin{bmatrix} \mathbf{T}_{0} &  \dots  &\mathbf{T}_{K}\end{bmatrix}$ and $\boldsymbol{\xi}=\begin{bmatrix} \boldsymbol{\xi}_{0} &  \dots  &\boldsymbol{\xi}_{K}\end{bmatrix}$.

To speed up optimization, we derive analytic Jacobian rather than inefficient automatic differentiation~\cite{dellenbach2022ct}. 
Indeed, each error term is only related to several poses. Our derivation uses the right forms in~\cite{sola2018micro} so that the concrete Jacobians of continuous-time registration term $\mathbf{e}_{_{k}\mathbf{z}_{i}^{j}} $ is 
\begin{subequations}
\begin{align}
    \frac{\partial \mathbf{e}_{_{k}\mathbf{z}_{i}^{j}}}{\partial \mathbf{s} } &=\frac{\partial \mathbf{e}_{_{k}\mathbf{z}_{i}^{j}}}{\partial \phi_{k}({_{k}t_{i}^{j}}) }
\begin{bmatrix}
  \frac{\partial \phi_{k}({_{k}t_{i}^{j}})}{\partial \mathbf{T}_{k-1} } &\frac{\partial \phi_{k}({_{k}t_{i}^{j}})}{\partial \mathbf{T}_{k} }  
\end{bmatrix}, \\
    \frac{\partial \mathbf{e}_{_{k}\mathbf{z}_{i}^{j}}}{\partial \phi_{k}({_{k}t_{i}^{j}}) }&=
{_{k}\mathbf{n}_{i}^{j}}^{\top }\begin{bmatrix}
  \mathbf{R}_{{_{k}t_{i}^{j}}} &
-\mathbf{R}_{{_{k}t_{i}^{j}}}\left [{\mathbf{T}^{B}_{L_{j}}}\cdot  {_{k}\mathbf{p}_{i}^{j}}  \right ]_{\times }
\end{bmatrix}, \\
    \frac{\partial \phi_{k}({_{k}t_{i}^{j}})}{\partial \mathbf{T}_{k-1} }&=
(1-{_{k}\alpha _{i}^{j}})\mathbf{J}_{r}(({_{k}\alpha _{i}^{j}}- 1)\boldsymbol{\tau}_{k})\mathbf{J}_{l}^{-1}(\boldsymbol{\tau}_{k}),\\
    \frac{\partial \phi_{k}({_{k}t_{i}^{j}})}{\partial \mathbf{T}_{k} }&={_{k}\alpha _{i}^{j}}\mathbf{J}_{r}({_{k}\alpha _{i}^{j}}\boldsymbol{\tau}_{k})\mathbf{J}_{r}^{-1}(\boldsymbol{\tau}_{k}).
\end{align}
\end{subequations}
$\mathbf{R}_{{_{k}t_{i}^{j}}}$ is the rotation part of reference coordinate pose $\mathbf{T}_{{_{k}t_{i}^{j}}}=\phi_{k}({_{k}t_{i}^{j}})$ at ${_{k}t_{i}^{j}}$. $\left [\cdot  \right ]_{\times }$ denotes the skew symmetric matrix. $\mathbf{J}_{r}(\cdot)$ and $\mathbf{J}_{l}(\cdot)$ are the right- and left- Jacobians~\cite{sola2018micro} of $\mathrm{SE}(3)$, respectively. The Jacobians of kinematic constraints are
\begin{equation}
             \frac{\partial \mathbf{e}_{v_{k}}}{\partial \mathbf{T}_{k-1}}=-\mathbf{J}_{l}^{-1}(\boldsymbol{\tau}_{k}),\quad
    \frac{\partial \mathbf{e}_{v_{k}}}{\partial \mathbf{T}_{k}}=\mathbf{J}_{r}^{-1}(\boldsymbol{\tau}_{k}).   
\end{equation}

\subsection{Map Management}\label{sec:map}
The bottleneck of LiDAR odometry is thousands of nearest-neighbor searches during point registration. In order to accelerate this intensive and repetitive operation, our map adopts a spatial hashing structure from~\cite{dellenbach2022ct,vizzo2023ral}. The voxel size depends on the environment. Each voxel stores up to 20 points and we explore the closest 7 voxels during nearest-neighbor searching. Generally, we assume the LiDAR odometry starts from a stationary state, and use the points collected within the first 0.3s for map initialization. The voxel index of each point is computed by its position in world coordinates. The map points are unchangeable during registration. Once the optimization is converged, the map will be updated by the points leaving the window. To reduce the memory consumption of the map, points located more than 100m away from the current LiDAR center are removed.

\section{EXPERIMENT}

In this section, we present the details of our experiments. To show the effectiveness and generalizability, we evaluate our approach named `Traj-LO' on three ordinary datasets and another with extreme motion where IMUs are saturated. The comparative methods include state-of-the-art LOs and LIOs. 

Practically, there exists a trade-off between maintaining a longer temporal window to cover a larger field of view (FoV) and ensuring computational efficiency. In our experiments, we choose 4 segments in the temporal window, with each interval $\Delta t_{k}$ at 0.03s. The noise $\mathbf{Q}_r = \sigma_{r}   \mathbf{I}_1$ and $\mathbf{Q}_v = \sigma_{v}   \mathbf{I}_6$ are diagonal matrices, where $\sigma_{r}=0.1$ and $\sigma_{v}=0.05$. The indoor voxel size is 0.4m and the outdoor is 0.8m.

\subsection{Dataset}

\textbf{KITTI odometry dataset}~\cite{geiger2012we} provides driving scenarios, in which 3D point scans are collected using the Velodyne HDL-64E S2. This dataset offers a wide range of scenarios from urban city to highway traffic. However, points in all 22 sequences are motion-corrected where temporal information is discarded. Although~\cite{geiger2013vision} provides raw LiDAR data, it does not retain the timestamps of individual points. Due to its popularity in the robotic community, we still report the evaluation results on this benchmark.

\textbf{NTU VIRAL}~\cite{nguyen2022ntu} is a visual-inertial-ranging-LiDAR dataset for autonomous aerial vehicles. It has two 16-channel Ouster LiDARs, which separately equip on the horizontal and vertical direction. These LiDARs produce point clouds at 10~Hz rate accompanied by a 9-axis IMU at 385~Hz. The point cloud contains the timestamp of each point relative to the scan's start time, which is crucial for continuous-time estimation. The ground truth is obtained by a Leica Nova MS60 in several challenging indoor and outdoor conditions.
 
\textbf{Hilti 2021 dataset}~\cite{helmberger2022hilti} contains indoor sequences of offices, labs, and construction environments and outdoor sequences of construction sites and parking areas. The LiDAR configuration is an Ouster OS0-64 which collected points in $360^{\circ }$ FoV at 10~Hz, and another LiDAR unit is Livox MID70 with a non-repetitive scan pattern in $70^{\circ }$ circular FoV at 10Hz. The devices are mounted on a handheld platform, providing millimeter-accurate ground truth from the Hilti PLT 300 automated Total Station or a MoCap system. The temporal information of each point is reserved. 

\textbf{Point-LIO dataset}~\cite{he2023point} is collected by Livox Avia. We select two sequences for evaluation. The outdoor sequence features a spinning motion on a rotating platform, while the indoor involves a circular swinging motion with the LiDAR attached to one end of a rope. Both sequences suffer that the kinematic state exceeds the IMU measuring range, causing most LIOs~\cite{xu2022fast,shan2020lio} to fail. The interval $\Delta t_{k}$ is 0.01s and voxel size is 0.2m for indoor sequence.

\begin{table}[t]
\centering

\caption{\small RTE Results (\%) on KITTI Odometry Benchmark.}
\label{tab:kitti}
\resizebox{0.5\textwidth}{!}{
\begin{threeparttable}
\begin{tabular}{@{}l|ccccccccccc|cc@{}}
\toprule
\midrule
Approach&00& 01& 02& 03&04&05&06&07&08&09&10 & AVG & Online\\
\midrule
FLOAM~\cite{wang2021f} &0.71&\textbf{0.71}&0.73&0.98&0.57&0.62&0.33&0.47&1.04&0.88&1.02& 0.73&0.72\\
KISS-ICP~\cite{vizzo2023ral} &0.52&0.72&0.53&\textbf{0.65}&\textbf{0.35}&0.30&\textbf{0.26}&0.33&\textbf{0.81}&0.49&0.54&\textbf{0.50}&0.61\\
CT-ICP~\cite{dellenbach2022ct} & \textbf{0.49}&0.76&\textbf{0.52}&0.72&0.39&\textbf{0.25}&0.27&0.31&\textbf{0.81}&0.49&\textbf{0.48}&\textbf{0.50}&0.59\\
Traj-LO (ours)&0.50&0.81&\textbf{0.52}&0.67&0.40&\textbf{0.25 }&0.27&\textbf{0.30}&\textbf{0.81}&\textbf{0.45}&0.55& \textbf{0.50}&\textbf{0.58}\\

\midrule
\bottomrule
\end{tabular}

\end{threeparttable}
}
\vspace{-0.2in}
\end{table}

\begin{table*}[t]
\vspace{-0.1in}
\centering
\caption{{\small ATE (m) on the NTU VIRAL Dataset.}}

\resizebox{\textwidth}{!}{
\begin{threeparttable}
\begin{tabular}{@{}llccccccccccccccccccc@{}}
\toprule
\midrule

 &\multirow{2}{*}{Approach}& \multirow{2}{*}{Sensor\tnote{1}} & \multicolumn{3}{@{}c@{}}{eee} & \multicolumn{3}{@{}c@{}}{nya}& \multicolumn{3}{@{}c@{}}{sbs}&\multicolumn{3}{@{}c@{}}{rtp} & \multicolumn{3}{@{}c@{}}{tnp}& \multicolumn{3}{@{}c@{}}{spms}\\

 \cmidrule(r){4-6}\cmidrule(r){7-9}\cmidrule(r){10-12}
\cmidrule(r){13-15}\cmidrule(r){16-18}\cmidrule(r){19-21}
 & & & 01&02&03 & 01&02&03 & 01&02&03 & 01&02&03 & 01&02&03 & 01&02&03\\
 \midrule
 \multirow{4}{*}{\rotatebox{90}{LO}}& FLOAM~\cite{wang2021f} &L1& 4.486&8.238&1.133&1.447&1.292&1.498&0.976&2.010&1.079&10.775&4.637&2.218&2.354&2.249&1.566&x&x&x\\
 & KISS-ICP~\cite{vizzo2023ral} & L1&2.220&1.570&1.014&0.628&1.500&1.272&0.917&1.312&1.030&3.663&1.970&2.382&2.305&2.405&0.799&8.493&x&5.451\\
   & CT-ICP~\cite{dellenbach2022ct} &L1& 7.763&0.125&11.171&0.100&0.101&0.073&x&0.084&1.545&x&0.081&0.086&0.073&0.071&\underline{0.045}&x&x&x\\

& MARS\tnote{3}~\cite{quenzel2021real} &L1+L2& 0.247&0.103&0.093&0.056&0.062&0.083&0.137&0.126&0.159&×&0.233&0.138&0.073&0.068&0.067&×&×&19.865  \\

 & \multirow{2}{*}{\bf{Ours}} &L1& 
0.055&0.039&\underline{0.035}&\underline{0.047}&0.052&0.050&0.048&\underline{0.039}&\underline{0.039}&0.050&\underline{0.058}&0.057&0.505&0.607&0.101&0.121&x&0.103\\

& &L1+L2&\underline{0.051}&\underline{0.033}&0.036&0.052&\underline{0.045}&\underline{0.045} & \underline{0.042}&0.042&0.041&\underline{0.045}&0.061&\underline{0.044}&\underline{0.049}&\underline{0.040}&0.049&\bf{\underline{0.056}}&0.163&\bf{\underline{0.063}}
\\

 \midrule
  \multirow{3}{*}{\rotatebox{90}{LIO}}&LIO-SAM~\cite{shan2020lio}&L1+I& 0.032&0.050&0.077&0.041&0.056&0.067&0.054&0.043&0.044&0.085&0.073&0.066&0.065&0.127&0.052&0.207&x&\underline{0.074}
  
  \\
   & FAST-LIO~\cite{xu2022fast} & L1+I& 
 \bf{\underline{0.029}}&\bf{\underline{0.019}}&\bf{\underline{0.023}}&0.031&0.031&0.036&0.031&0.045&\bf{\underline{0.029}}&\bf{\underline{0.042}}&0.060&\bf{\underline{0.052}}&0.043&0.037&0.045&\underline{\bf{0.056}}&\bf{\underline{0.050}}&0.075
 \\

  & SLICT\tnote{3}~\cite{nguyen2023slict} &L1+L2+I&0.032&0.025&0.028&\bf{\underline{0.023}}&\bf{\underline{0.023}}&\bf{\underline{0.016}}&\bf{\underline{0.030}}&\bf{\underline{0.029}}&0.034&0.045&\bf{\underline{0.047}}&0.050&\bf{\underline{0.029}}&\bf{\underline{0.020}}&\bf{\underline{0.038}}&0.061&0.10&0.066
 \\

  &CLIC\tnote{3}~\cite{lv2023continuous}
&L1+L2+I&0.040&0.021&0.031&0.030&0.037&0.034&0.033&0.037&0.044&0.072&0.239&0.064&0.060&0.061&0.053&0.123&x&0.211\\

\midrule
\bottomrule
\end{tabular}

\begin{tablenotes}
\item[1] L1: horizon OS1-16, L2: vertical OS1-16, I: external IMU. 
\item[2] The best results overall are in \textbf{blod}, while the best results in each category are \underline{underlined}. 'x' denotes divergence.
\item[3] The results of MARS and SLICT are obtained from the paper~\cite{nguyen2023slict} while CLIC are from~\cite{lv2023continuous}. 

\end{tablenotes}
\end{threeparttable}
}
\label{tab:ntu}
\vspace{-0.2in}
\end{table*}


\begin{table}[t]
\centering
\caption{{\small ATE (m) on the Hilti SLAM Challenge Dataset}}
\resizebox{0.5\textwidth}{!}{
\begin{threeparttable}
\begin{tabular}{@{}llccccccc@{}}
\toprule
\midrule
& Approach& Sensor\tnote{1}& RPG 
  & Base1
 & Base4 & Lab
 & Cons2 & Camp2  \\
\midrule
 \multirow{8}{*}{\rotatebox{90}{LO}}& \multirow{2}{*}{FLOAM~\cite{wang2021f}} &L1& 2.775&0.914&0.287&0.182&11.515& 8.946\\
 & & L2& - &  - &- & - & - & -\\
 & \multirow{2}{*}{KISS-ICP~\cite{vizzo2023ral}} & L1&0.187&\bf{\underline{0.294}}&0.119&0.073&0.835&5.052\\
 & & L2 &3.726&6.850 &0.293&×&21.650&3.609\\
 & \multirow{2}{*}{CT-ICP~\cite{dellenbach2022ct}}  & L1 &0.188&0.321&0.210&0.052&0.087&0.077\\
& & L2 & 0.197&×&0.126&×& 13.359&6.516\\

 & \multirow{3}{*}{\bf{Ours}} &L1& 0.172 &  0.301&0.064& \bf{\underline{0.027}}&0.065&\bf{\underline{0.045}}\\
 &&L2& 0.218&0.314&0.064&×&0.135&0.049\\
   & &L1+L2 & \bf{\underline{0.170}}&0.297&\underline{0.049}& 0.050&\bf{\underline{0.063}}&0.060\\

 \midrule
  \multirow{3}{*}{\rotatebox{90}{LIO}}
  &LIO-SAM~\cite{shan2020lio}& -&-&-&-&-&-&- \\
 & \multirow{2}{*}{FAST-LIO~\cite{xu2022fast}} & L1+I&\underline{0.182}&\underline{0.309}&\bf{\underline{0.033}}&\underline{0.035}&\underline{0.066}&\underline{0.087}\\
 & & L2+I&0.282&0.313&0.693&×&0.198&0.063\\

   & \multirow{2}{*}{CLIC~\cite{lv2023continuous}}  & L1+I&0.394& 0.340&0.202&0.240&0.327&0.397\\
  & & L2+I& -&-&-&-&-&- \\

\midrule
\bottomrule
\end{tabular}

\begin{tablenotes}
\item[1] L1: OSO-64, L2: Livox MID70, I: IMU embedded in L1.
\item[2] The best results overall are in \textbf{blod}, while the best results in each category are \underline{underlined}. 'x' denotes divergence, and '-' denotes invalid results.
\end{tablenotes}

\end{threeparttable}
}
\label{tab:hilit}
\vspace{-0.2in}
\end{table}

\subsection{Comparison With LiDAR-only Methods}
We compare Traj-LO with the LiDAR-only methods, including FLOAM~\cite{wang2021f}, KISS-ICP~\cite{vizzo2023ral} and CT-ICP~\cite{dellenbach2022ct}. 

\subsubsection{KITTI}
At first, a vertical angle of $0.205^{\circ }$ is used to rectify the calibration errors in raw point clouds. Since the points are motion-corrected, continuous registration is disabled. We maintain motion constraints between 4 consecutive scans. For quantitative analysis, we use the KITTI relative translation error, and the results are reported in Table~\ref{tab:kitti}. It can be seen that FLOAM obtains the worst performance among the four methods. Across the 11 training sequences, KISS-ICP, CT-ICP and our Traj-LO achieve similar average accuracy. On the other 11 testing sequences, we have obtained the best online result on the KITTI benchmark with a score of 0.58\% translation error and 0.0014deg/m rotation error.

\subsubsection{NTU}

As a pioneering benchmark that leads the development of LiDAR Odometry in the last decade, KITTI mostly moves straightforward plus small turns. This cannot fulfill the requirement of robust LiDAR odometry nowadays. Thus, we provide a comprehensive evaluation on more challenging aerial dataset. It records the specific timestamps of each LiDAR point, which makes it possible to examine the advantage of continuous-time approaches. The results of absolute trajectory error (ATE) compared with ground truth are listed in Table~\ref{tab:ntu}, which is computed by evo  package\footnote{https://github.com/MichaelGrupp/evo}. 

Our proposed Traj-LO approach achieves the best performance and outperforms other methods at a large margin except for tnp sequences. FLOAM nearly drifts on all sequences, since its original public code does not compensate for motion distortion. Although KISS-ICP claims its constant velocity model is on par or even slightly better with the IMU, it drifts on most sequences. For the more complicated NTU dataset, the constant velocity assumption may not be valid.

CT-ICP achieves an average accuracy at the centimeter level, similar to our Traj-LO when excluding failed sequences, which outperforms the two previous discrete-time methods. The robustness of CT-ICP is unsatisfactory, as it fails on half of the NTU sequences. This is because linear interpolation using the beginning and end poses of the scan is unsuitable in the case of rapid motion. Besides, it is not enough to track the aerial vehicles by only taking consideration of the translation in motion constraint.

\subsection{Comparison With LiDAR-inertial Methods}
Traj-LO has demonstrated a significant advantage over the existing LiDAR-only methods. Thus, we also compare it against the widely used LIO methods like LIO-SAM~\cite{shan2020lio} and FAST-LIO~\cite{xu2022fast}. Surprisingly, Traj-LO performs the best on the handheld Hilti dataset in Table~\ref{tab:hilit} and achieves the competitive results on the aerial NTU dataset in Table~\ref{tab:ntu}. Note that our approach achieves such promising results using only LiDAR points.

\begin{figure}[t]
	\centering
\includegraphics[width=0.45\textwidth]{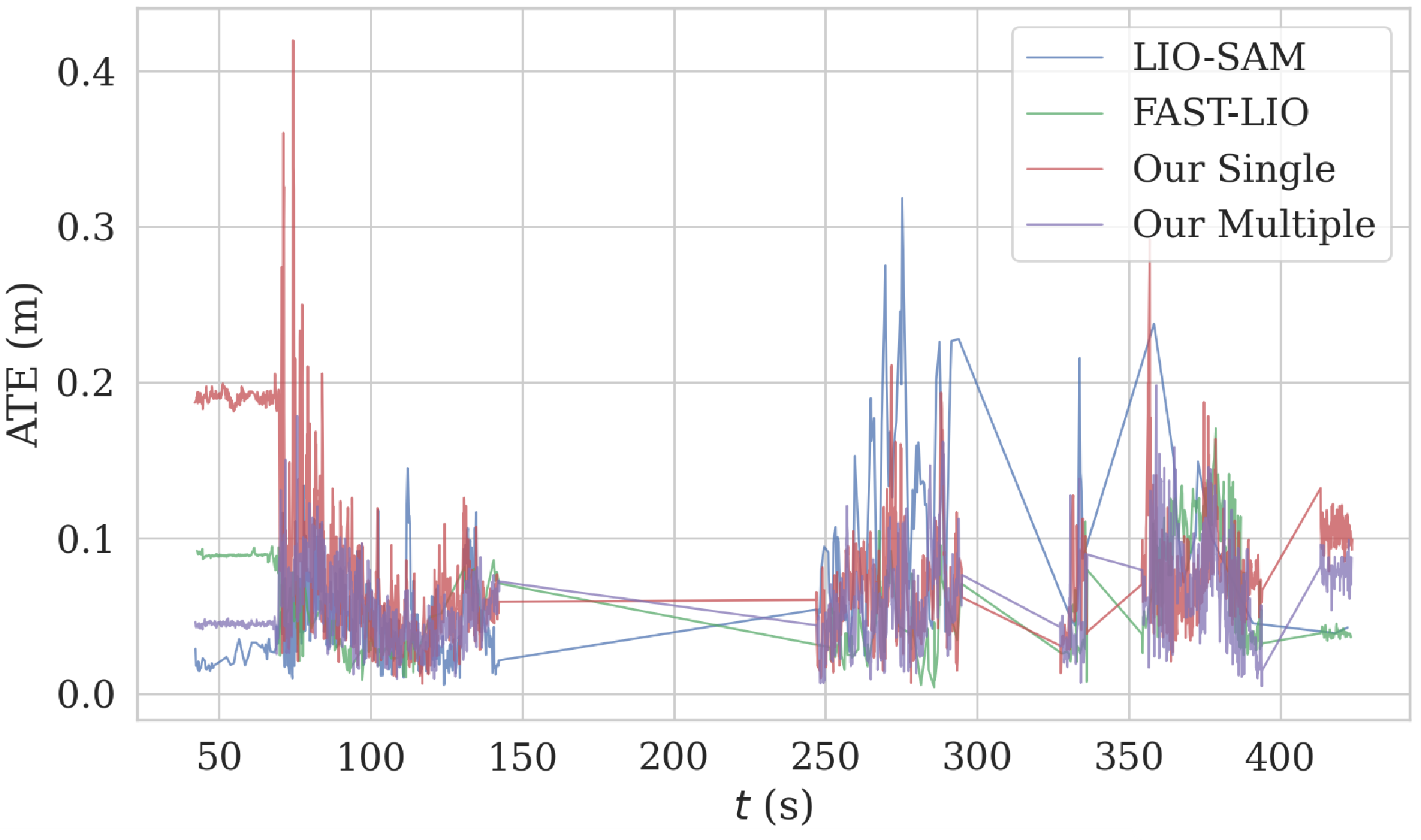}
	\caption{\small \textit{ The ATE of four methods on sequence spms\_03 is depicted over time. The notable differences occur mainly during the take-off and landing phases. }}
	\label{fig:spms03}
	\vspace{-0.3in}
\end{figure}

The difficulty of the NTU dataset lies in the jerky motion patterns during take-off and landing. Fig.~\ref{fig:spms03} plots the ATE of spms\_03 over time. For a single horizontal LiDAR, Traj-LO exhibits a larger ATE at the beginning and end of the trajectory while its ATE closely matches that of LIO-SAM and FAST-LIO during smooth flight. This indicates that the IMU is able to provide the extra kinematics information when lacking sufficient geometric constraints.

\subsection{Evaluation on Multi-LiDAR System}
\begin{figure}[t]
	\centering
	\includegraphics[width=0.48\textwidth]{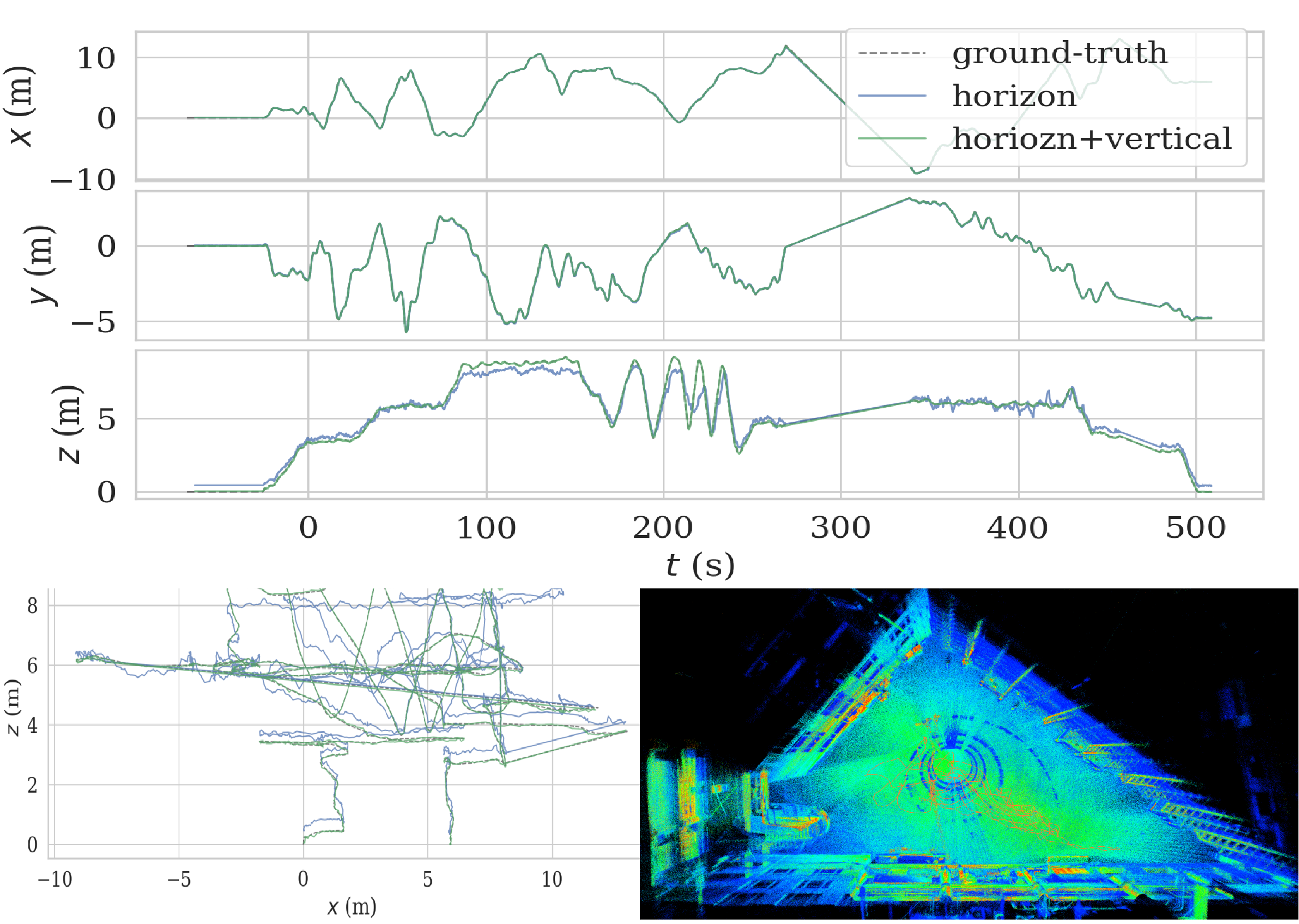}
	\caption{\small \textit{ The environment in the tnp sequence resembles a Manhattan world, with nearly all 16-channel laser beams from the horizontal LiDAR sensor interacting with three vertical walls. The upper figure provides separate plots of the XYZ positions of tnp\_01 over time, while the bottom left plots the trajectory in the xz plane. The bottom right shows the mapping results after fusing the points from both the horizontal and vertical LiDAR.}}
	\label{fig:degrade}
	\vspace{-0.3in}
\end{figure}

Our continuous-time trajectory representation is naturally capable of integrating multiple LiDARs into a unified framework. As demonstrated in Table~\ref{tab:ntu}, Traj-LO outperforms the existing B-spline-based LO method MARS~\cite{quenzel2021real}, when utilizing both horizontal and vertical LiDARs. Furthermore, our approach exhibits exceptional performance in high-speed spms sequences that is traditionally considered as a strength of LIO approaches such as CLIC~\cite{lv2023continuous} and SLICT~\cite{nguyen2023slict}.

A notable advantage of multi-LiDAR systems is their ability to provide supplementary geometric information in scenes, where a single LiDAR may degrade. As illustrated in Fig.~\ref{fig:degrade}, tnp sequences lack sufficient features to constrain motion in the vertical direction. CT-ICP with single LiDAR maintains low ATE by searching more neighborhoods from the 27 closest voxels. In contrast, Traj-LO leverages the additional vertical LiDAR to enhance system observability, which significantly reduces large errors along the z-direction.

As shown in Fig.~\ref{fig:spms03}, the ATE of Traj-LO during takeoff and landing phases is substantially reduced by introducing the vertical LiDAR. This indicates us that the LiDAR-only method is able to achieve similar results as LIO through providing the adequate geometric constraints for registration.

\subsection{Evaluation on Non-repetitive LiDAR}
Non-repetitive scanning LiDARs with a small FoV are popular as complements to multi-line spinning LiDARs. In the experiment, we evaluate different methods on the Hilti dataset with a Livox MID70. To cover the $70^{\circ}$ circular FoV with limited laser heads, the laser direction has to change frequently using Risley prism. This sensor introduces more severe motion distortion~\cite{zheng2023ectlo}. As shown in Table~\ref{tab:hilit}, continuous-time methods outperform discrete-time methods even without IMU. However, the overall performance of Livox MID70 is lower than that of multi-line Ouster LiDAR, which offers a wider FoV.

\subsection{Extreme Scenarios Beyond IMU Measuring Range}

\begin{figure}[t]
	\centering
	\includegraphics[width=0.44\textwidth]{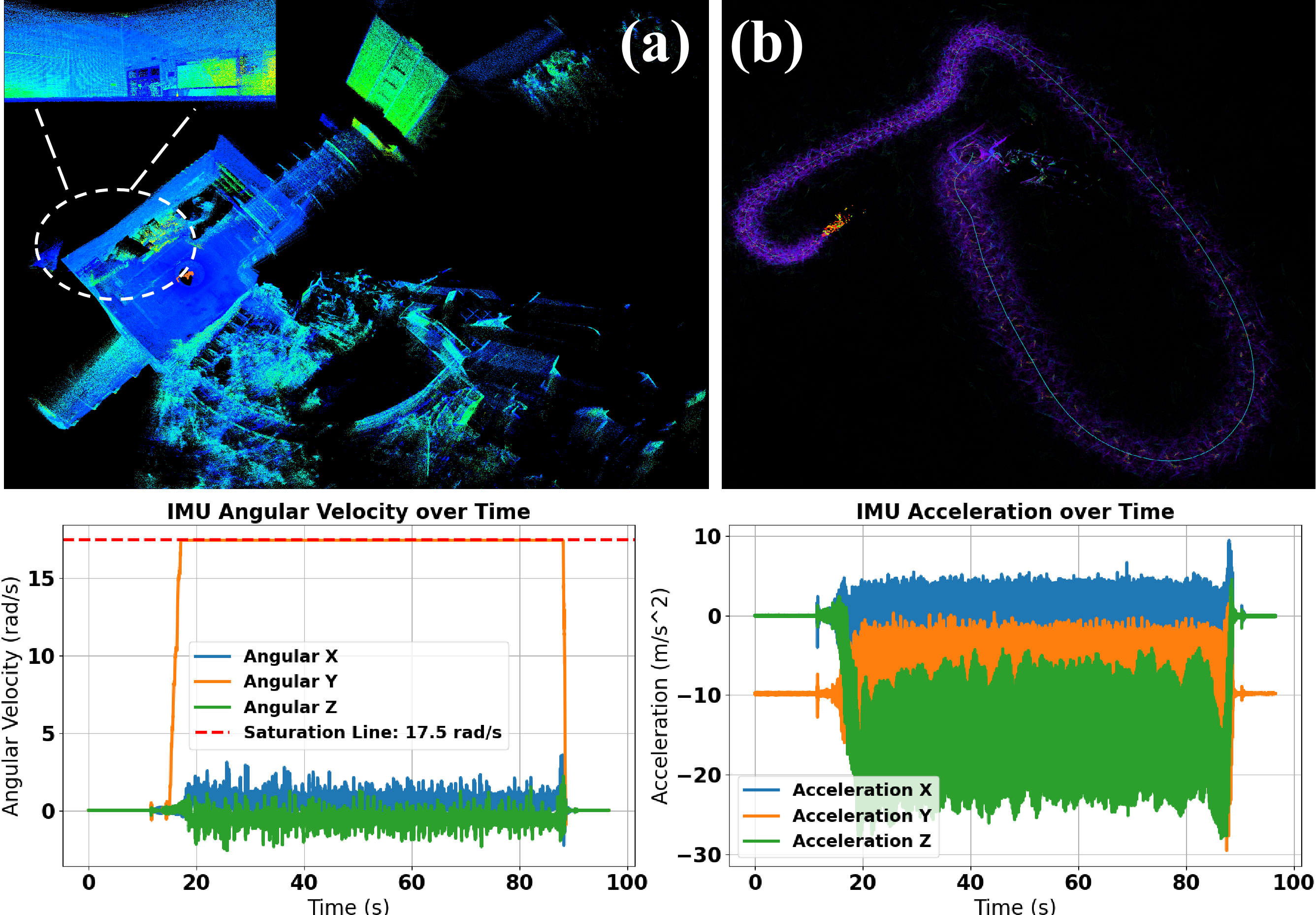}
	\caption{\small \textit{ The bottom shows the IMU data collected by the rotating platform, where angular velocity exceeds the measurement range of 17.5 rad/s. (a) is the mapping result of our LiDAR-only method while (b) FAST-LIO~\cite{xu2022fast} drifts significantly.}}
	\label{fig:outdoor}
	\vspace{-0.3in}
\end{figure}

Finally, we evaluate Traj-LO in more challenging scenarios using the Point-LIO dataset. Fig.~\ref{fig:extreme} and Fig.~\ref{fig:outdoor} show the mapping results along with the recorded IMU data over time. Remarkably, our LiDAR-only odometry method exhibits minimal drift even in case of exceptionally aggressive motion. This unexpected performance can be attributed to the spatial-temporal information contained within the millions of streaming LiDAR points, which surpasses the capabilities of conventional IMUs with the frequencies between 100-400~Hz. Furthermore, MEMS IMUs tend to introduce noises, especially in aggressive motions, whereas the range measurements obtained from individual LiDAR points are considerably more accurate. By making use of previously overlooked temporal data in registration, LiDAR-only odometry demonstrates a level of capability that exceeds the initial expectations.

\section{CONCLUSION}~\label{sec:conc}
This letter has introduced a LiDAR-only odometry using a simple yet effective continuous-time trajectory. By coupling geometric information and kinematic prior hiding behind streaming points, the performance of our approach was on par with start-of-the-art LIOs and even suppressed them in extreme scenarios. Moreover, the proposed odometry was designed to accommodate various types of LiDARs as well as multi-LiDAR systems. At present, our method is primarily designed for odometry task. In future work, we intend to explore the capabilities of a complete LiDAR-only SLAM system that incorporates this continuous-time trajectory along with a more representative map structure.

\bibliographystyle{IEEEtran}
\bibliography{mybib}

\end{document}